\documentclass{acm_proc_article-sp}

\newfont{\mycrnotice}{ptmr8t at 7pt}
\newfont{\myconfname}{ptmri8t at 7pt}
\let\crnotice\mycrnotice%
\let\confname\myconfname%

\usepackage{amsmath,amssymb}
\usepackage{pstricks}
\usepackage{graphicx}
\usepackage{xspace}            
\usepackage{graphicx}
\usepackage{multirow}
\usepackage{subfig}
\usepackage{multirow}
\usepackage{array}
\usepackage{url}
\usepackage{pdfpages}
\usepackage{booktabs}
\usepackage{balance}
\usepackage[sort&compress,square,numbers,comma]{natbib}

\newcommand{\vct}[1]{\ensuremath{\boldsymbol{#1}}}
\newcommand{\mat}[1]{\ensuremath{\mathtt{#1}}}
\newcommand{\set}[1]{\ensuremath{\mathcal{#1}}}
\newcommand{\con}[1]{\ensuremath{\mathsf{#1}}}
\newcommand{\T}{\ensuremath{^\top}}

\newcommand{\Malheur}{\texttt{Malheur}\xspace}

\newcommand{\ie}{{i.e.}\xspace}
\newcommand{\eg}{{e.g.}\xspace}
\newcommand{\etal}{{et al.}\xspace}

\newcommand{\ourparagraph}[1]{\smallskip \noindent \textbf{#1.}}

\newcommand{\comment}[1]{}

\permission{Preprint of the work published at AISec 2014. Please cite as: B. Biggio, K. Rieck, D. Ariu, C. Wressnegger, I. Corona, G. Giacinto, and F. Roli. Poisoning behavioral malware clustering. In 2014 Workshop on Artificial Intelligent and Security, AISec '14, pages 27-36, New York, NY, USA, 2014. ACM.}

\begin{document}
\title{Poisoning Behavioral Malware Clustering}

\numberofauthors{7}
\author{
\alignauthor Battista Biggio\\
       \affaddr{Universit\`a di Cagliari}\\
       \affaddr{Piazza d'Armi}\\
       \affaddr{09123, Cagliari, Italy}\\
       \email{battista.biggio@diee.unica.it}
\alignauthor Konrad Rieck\\
       \affaddr{University of G\"{o}ttingen}\\
       \affaddr{Goldschmidtstra{\ss}e 7}\\
       \affaddr{37077, G\"{o}ttingen, Germany}\\
       \email{konrad.rieck@uni-goettingen.de}
	\and
\alignauthor Davide Ariu\\\
       \affaddr{Universit\`a di  Cagliari}\\
       \affaddr{Piazza d'Armi}\\
       \affaddr{09123, Cagliari, Italy}\\
       \email{davide.ariu@diee.unica.it} 
\alignauthor Christian Wressnegger\\
       \affaddr{University of G\"{o}ttingen}\\
       \affaddr{Goldschmidtstra{\ss}e 7}\\
       \affaddr{37077, G\"{o}ttingen, Germany}\\
       \email{christian.wressnegger@\\cs.uni-goettingen.de}
\alignauthor Igino Corona\\
       \affaddr{Universit\`a di  Cagliari}\\
       \affaddr{Piazza d'Armi}\\
       \affaddr{09123, Cagliari, Italy}\\
       \email{igino.corona@diee.unica.it}
       \and 
\alignauthor Giorgio Giacinto\\
       \affaddr{Universit\`a di Cagliari}\\
       \affaddr{Piazza d'Armi}\\
       \affaddr{09123, Cagliari, Italy}\\
       \email{giacinto@diee.unica.it}
\alignauthor Fabio Roli\\
       \affaddr{Universit\`a di Cagliari}\\
       \affaddr{Piazza d'Armi}\\
       \affaddr{09123, Cagliari, Italy}\\
       \email{roli@diee.unica.it}
}

\maketitle              

\begin{abstract}
Clustering algorithms have become a popular tool in computer security to analyze the behavior of malware variants,  identify novel malware families, and generate signatures for antivirus systems.
However, the suitability of clustering algorithms for security-sensitive settings has been recently questioned by showing that they can be significantly compromised if an attacker can exercise some control over the input data.
In this paper, we revisit this problem by focusing on behavioral malware clustering approaches, and investigate whether and to what extent an attacker may be able to subvert these approaches through a careful injection of  samples with poisoning behavior. 
To this end, we present a case study on Malheur, an open-source tool for behavioral malware clustering. Our experiments not only demonstrate that this tool is vulnerable to poisoning attacks, but also that it can be significantly compromised even if the attacker can only inject a very small percentage of attacks into the input data. As a remedy, we discuss possible countermeasures and highlight the need for more secure clustering algorithms.
\end{abstract}




\section{Introduction}
\label{sect:introduction}
Automated techniques for behavioral clustering of malware have been found to be effective for the development of analysis, detection and mitigation strategies against a broad spectrum of malicious software. Such techniques can significantly ease the identification of polymorphic instances of well-known malware as well as novel attack types and infection strategies, reducing by orders of magnitude the burden of the analysis task~\cite[e.g.,][]{RieHolWilDueLas08,rieck11-jcs,JanBruVen11,perdisci13}. 

Behavioral clustering is motivated by a key assumption: albeit malware writers can generate a large number of polymorphic variants of the same malware, \eg, using executable packing and other code obfuscation techniques~\cite{Guo08, fogla06}, these polymorphic variants will eventually perform similar activities when executed.
To expose these behavioral similarities, malware binaries are usually executed in a monitored sandbox environment, in order to identify malware families characterized by similar host-level events~\citep[\eg,][]{BaiObeAndMaoJahNaz07, RieHolWilDueLas08, BayComHlaKruKir09, JanBruVen11} or network traffic patterns~\citep[\eg,][]{GuPerZhaLee08,GuZhaLee08,PerLeeFea10,perdisci13}. 

However, regardless the behavioral features being used, all these proposals suffer from the same vulnerability: \emph{clustering algorithms have not been originally devised to deal with data from an adversary}.
As outlined in recent work~\cite{biggio13-aisec,biggio14-spr}, this may allow an attacker to devise carefully-crafted attacks that can significantly compromise the clustering process itself, and invalidate subsequent analyses.

In this work, we also show that the effectiveness of clustering algorithms --- in particular, single-linkage clustering --- can be dramatically reduced by a skilled adversary through a proper, deliberate manipulation of malware samples, in the context of a more realistic application scenario that those considered in~\cite{biggio13-aisec,biggio14-spr}. 
To this end, we first review the attacker's model proposed in~\cite{biggio13-aisec,biggio14-spr}, as it can also be exploited as a general threat model for behavioral malware clustering, and then investigate a worst-case attack against \Malheur{}~\cite{rieck11-jcs}, an open-source malware clustering tool. We emulate an attacker who \emph{adds} specially-crafted poisoning actions to the original behavior of malware samples, thus leaving intact their original malicious goals. Our experimental results clearly show that even a small fraction of 3\% of poisoning samples may completely subvert the clustering process, leading to poor clustering results. Thus, our case study highlights the need for \emph{robust} malware clustering techniques, capable of coping with malicious noise. As a consequence, a safe application of clustering algorithms for malware analysis remains an open research issue. Throughout the paper we sketch some promising ways of research towards this goal.

\ourparagraph{Contributions} In summary, the main contribution of this paper is to
 extend and adapt the poisoning attacks proposed in \cite{biggio13-aisec,biggio14-spr} against the single-linkage hierarchical clustering algorithm to target \Malheur{}~\cite{rieck11-jcs}, an open-source tool for behavioral malware clustering. In this case, the main difficulty with respect to previous work relies in constructing real malware samples that correspond to the desired, optimal feature vectors found by the optimal attack strategy, while accounting for application-specific constraints on the manipulation of the feature values of each sample. This is a well-known issue in the field of adversarial machine learning, referred to as the problem of \emph{inverting} the feature mapping~\cite{biggio14-tkde,huang11}. 
To assess the effectiveness of poisoning attacks against behavioral malware clustering, we finally report an extensive set of experiments that highlight the vulnerability of such  approaches to well-crafted attacks, as well as the need for identifying suitable countermeasures, for which we identify some interesting ways of research.

\ourparagraph{Organization}
The remainder of this paper is structured as follows. In Sect.~\ref{sect:behavclust}, we give an overview of recent work on behavioral malware clustering. The previously-proposed framework for the security evaluation of clustering algorithms~\cite{biggio13-aisec,biggio14-spr} is discussed in Sect.~\ref{sect:attacking-clustering}. In Sect.~\ref{sect:poisoning-attacks}, we review the derivation of (worst-case) \emph{poisoning} attacks, in which the attacker has perfect knowledge of the targeted system. In Sect.~\ref{sect:malheur}, we describe \Malheur{}, the malware clustering tool exploited as a case study to evaluate our poisoning attacks. The latter are defined as variants of the previously-proposed poisoning attacks, to deal with the specific feature representation exploited by \Malheur{}, in Sect.~\ref{sect:evaluation}, where we also report the results of our experimental evaluation. Conclusions are discussed in Sect.~\ref{sect:conclusions}, along with possible future research directions.

\section{Malware Clustering}
\label{sect:behavclust}

The urgent need for automated analysis of malware naturally comes with the
ever-growing number of malicious codes on the Internet. In recent years,
machine learning techniques have received  attention in this area, as they
 enable improving the automation  of malware analysis. One prominent
representative are clustering algorithms. These algorithms enable grouping
similar malware automatically and can thereby reduce the  manual efforts
required for developing mitigation and detection techniques. Several
approaches for such a clustering have been devised in the last years, most notably, (a)~\emph{clustering of network traffic}, and
(b) \emph{clustering of program behavior}.

\ourparagraph{Clustering of network traffic} Network communication is a  key
component of  malware and thus several malware families can be solely
characterized by their network traffic. For example, Gu \etal correlate
spatial-temporal relations in botnet communication using
clustering~\cite{GuZhaLee08}. To this end, the authors make use of
hierarchical clustering  on the basis of $q$-grams over a so-called
``activity log'' which describes a botnet's network communication in terms of
different types of responses. This approach is then extended to a more
general concept of C\&C communication~\cite{GuPerZhaLee08}, where 
the authors attempt to be agnostic to the protocol used as well as
the concrete hierarchy of the botnet.

In a similar line of research, Perdisci \etal~\cite{PerLeeFea10} focus on
HTTP-based malware with the objective to automatically generate network
signatures for malware. In particular, they use single-linkage clustering
over three stages, mainly to reduce computational complexity: first, a ``coarse-grained'' clustering is performed;
each of the corresponding clusters is then subdivided into 
a more  ``fine-grained'' set of clusters; and, eventually, similar clusters are merged together to avoid redundant signature generation.
An extension by the same authors~\cite{perdisci13} focuses more on the scalability of the proposed approach, in terms of the number of samples that the system is able to process in a given amount of time (\ie, the so-called \emph{throughput}).
The authors utilize an approximate clustering algorithm for the first stage of
their approach. This not only speeds up the initial stage but also decreases the
need of a merging phase, thus yielding a significant increase of the overall throughput of the system.

\ourparagraph{Clustering of program behavior} A second strain of
research has considered  program behavior of malware for
identifying related samples.  Despite polymorphism and obfuscation,
variants of the same malware family often show similar program
behavior. Bailey \etal~\cite{BaiObeAndMaoJahNaz07} have been the
first to apply clustering algorithms to this information. In
particular, they obtain a single-linkage clustering by computing
pairwise distances between sequences of host-level events. This
approach, however, has a quadratic runtime complexity and therefore
quickly reaches its limits in terms of the possible throughput.

Bayer \etal~\cite{BayComHlaKruKir09} counter this shortcoming with an approximate  clustering using
locality sensitive hashing (LSH). This makes it possible to scale the
analysis to several thousand malware samples. The behavioral analysis is
powered by the  malware analysis system Anubis~\cite{website:anubis}. Closely
related to this approach is the tool \Malheur~\cite{rieck11-jcs}, which we use
in our case study to demonstrate the effectiveness of our attacks. \Malheur
makes use of program behavior monitored by CWSandbox  in MIST
Format~\cite{WilHolFrei07,TriWilHolRie10} and is described in more detail in
Sect.~\ref{sect:malheur}.

More recently, several extensions have been proposed for improving behavioral
clustering of malware in practice. For example, Jang
\etal~\citep{JanBruVen11} apply feature hashing for clustering large sets of
malware binaries, Perdisci \& U~\citep{Perdisci12} propose an automatic
procedure for calibrating  clustering algorithms, and Hu \&
Shi~\citep{HuShi13} combine behavioral clustering with static code
analysis.

\smallskip
Although each of the presented approaches provides advantages for
keeping abreast of malware development, all approaches employ standard
clustering algorithms which have not been originally designed to explicitly cope with malicious noise. Consequently, the attacks proposed in this paper can be potentially adapted to several of these approaches with minor modifications.

\section{Security Evaluation of\\Clustering Algorithms}
\label{sect:attacking-clustering}

In this section we briefly review the framework proposed by Biggio~\etal~\cite{biggio13-aisec,biggio14-spr} for the security evaluation of  \emph{unsupervised} learning algorithms (including clustering) against adversarial attacks.
Similarly to previous work on the security evaluation of \emph{supervised} learning algorithms~\cite{biggio14-tkde,huang11,barreno-ASIACCS06}, this framework relies on a threat model that consists of defining the adversary's goal, knowledge of the attacked system, and capability of manipulating the input data, in order to formalize an \emph{optimal} attack strategy.

In the sequel, we describe this framework using the same notation defined in Biggio~\etal~\cite{biggio13-aisec,biggio14-spr}. We refer to any clustering algorithm as a function $f$ that maps a given dataset $\mathcal D=\{\vct x_i\}_{i=1}^{\con n}$ to a clustering result $\set C=f(\set D)$, without specifying the structure of $\set C$ at this stage, as it depends on the given clustering algorithm.

\subsection{Adversary's Goal}
\label{sect:adversary-goal}
The adversary's goal can be defined in terms of the desired security violation, and of the so-called attack specificity~\cite{biggio13-aisec,biggio14-spr,biggio14-tkde,huang11,barreno-ASIACCS06}.
A security violation may compromise the system \emph{integrity}, its \emph{availability}, or the \emph{privacy} of its users. \emph{Integrity violations}, in general, aim to perform some malicious activity without compromising the normal system operation. In the unsupervised learning setting, they have thus been defined as attacks aimed at changing the clustering of a given set of samples, without significantly altering the clustering result on the rest of the data.
\emph{Availability violations} aim to compromise system operation, causing a Denial of Service (DoS).
Therefore, in the unsupervised setting, availability attacks have been defined as attacks that aim to subvert the clustering process by altering its result as much as possible.
By contrast, \emph{privacy violations} are defined as attacks that may allow the attacker to gather information about the system's users by reverse-engineering the clustering process.
The attack specificity can be \emph{targeted} or \emph{indiscriminate}, depending on whether the attack aims to modify the clustering output only on a specific subset of samples, or indiscriminately on any sample.

\subsection{Adversary's Knowledge}
\label{sect:adversary-knowledge}
In order to achieve her goal, the adversary may exploit information at different abstraction levels about the targeted system. We summarize them in the following. First, the attacker may know the whole dataset $\set D$, a subset of it, or more realistically, only a \emph{surrogate} dataset $\set S$, that might be obtained from the same source of $\set D$, \eg, publicly available malware blacklists. Second, the adversary might be aware of, and reproduce, the extraction process of the whole feature set, or a portion of it. Indeed, when it comes to attacking open-source tools such as \Malheur , the adversary clearly has full knowledge of the feature set. 
Finally, the adversary might be aware of the targeted clustering algorithm, as well as of its initialization parameters (if any). In the case of \Malheur , this translates into knowing the user-specified configuration of the tool.

\ourparagraph{Perfect knowledge} The worst-case scenario in which the attacker has full knowledge of the targeted system is usually referred to as \emph{perfect knowledge}~\cite{biggio14-tkde,biggio13-ecml,biggio14-svm-chapter,biggio12-icml,kloft10,bruckner12,huang11,barreno-ASIACCS06}.
In our case, this amounts to knowing the data, the feature space, the clustering algorithm and its initialization (if any).

\subsection{Adversary's Capability}
\label{sect:adversary-capability}

The adversary's capability specifies how and to what extent the adversary can manipulate the input data to alter the clustering process. In several cases it is realistic to consider that the attacker can add a maximum number of (potentially manipulated) samples to the dataset $\set D$, without affecting the rest of the data.
For instance, anyone, including a skilled adversary, can submit novel malware samples to publicly-available malware-analysis services such as VirusTotal~\cite{VirusTotal} and Anubis~\cite{website:anubis}, which can in turn be used as sources to collect malware by registered users. If malware is collected from them, and clustered afterwards, the adversary may actually control a (small) percentage of the input data given to the clustering algorithm.

An additional constraint may be given in terms of how malware samples can be manipulated. In fact, to preserve its malicious functionality, malware code may not be manipulated in an unconstrained manner. Such a constraint can be often encoded by a suitable distance measure between the original, non-manipulated attack samples and the manipulated ones, as in~\cite{biggio14-tkde,kolcz09,huang11,barreno-ASIACCS06}. However, this strictly depends on the specific application and feature representation.

\subsection{Attack Strategy} 
\label{sect:attack-strategy}

Based on the presented threat model, consisting of assumptions on the adversary's goal, knowledge and capabilities, we can finally define the \emph{optimal} strategy for attacking a clustering algorithm as:
\begin{equation}
	\displaystyle
	\begin{array}{rl}
		\text{maximize}&\mathbb E_{\theta\sim\mu}[g(\set A';\theta)]\\
		\text{s.t.}&\set A'\in\Omega(\set A)\, .
	\end{array}
	\label{eq:optim}
\end{equation}
In this formulation, as in~\cite{biggio13-aisec,biggio14-spr}, the adversary's knowledge is characterized by a parameter vector $\theta$, whose elements embed information about the input data $\set D$, the clustering algorithm $f$, and its parameters (as discussed in Sect.~\ref{sect:adversary-knowledge}). The uncertainty of the adversary about the elements of $\theta$ is captured by a probability distribution $\mu$ defined over the set of all possible configurations $\theta$. 
Moreover, the objective function $g(\set A';\theta)\in\mathbb R$ measures the extent to which the adversary's goal is fulfilled by the set of attack samples $\set A'$ used to taint the initial data $\set D$, given the knowledge $\theta$.
In the above formulation, we consider the maximization of the expected value of this function with respect to $\theta$ sampled from the distribution $\mu$, denoted as $\mathbb E_{\theta\sim\mu}[\cdot]$.
Finally, the adversary's capability is encoded by the set $\Omega(\set A)$, which denotes the possible manipulations that the attacker can make on a given a set of attack samples $\set A$ before adding them to the original set $\set D$. The set $\set A$ of initial attacks can be empty, \eg, if the attack samples can be generated from scratch without preserving or exhibiting any malicious functionality.

It is finally worth remarking that the above optimization problem is formulated in terms of the considered feature representation, as many other adversarial machine learning problems~\cite{biggio13-aisec,biggio14-tkde,huang11,biggio13-ecml,biggio14-svm-chapter}.
In practice, after solving this problem, we are given a set of \emph{optimal feature vectors} for which we have to subsequently build a set of corresponding \emph{real samples}  to practically execute the attack.
This is clearly an application-specific problem that may not be trivial to solve depending on the given feature representation.
However, it can be mitigated by incorporating specific constraints on the manipulation of the feature values of the attack samples, while defining the set $\Omega$, as we will see in the next sections.

\section{Poisoning Attacks with\\Perfect Knowledge}
\label{sect:poisoning-attacks}

Following the framework described in the previous section, \emph{poisoning attacks} are defined as indiscriminate availability violations (\ie, DoS attacks) in which the attacker aims to maximally alter the clustering result on any of the input samples through the injection of well-crafted \emph{poisoning} samples. In the case of malware clustering, this amounts to adding carefully-designed malware samples to the input data to avoid the correct clustering of malware exhibiting similar behavior and, thus, the correct identification of both known and novel malware families.

As in previous work \cite{biggio13-aisec,biggio14-spr}, we are interested in analyzing the worst possible performance degradation that the system may incur under this attack. We therefore assume that the attacker has perfect knowledge of the targeted system, as described in Section~\ref{sect:adversary-knowledge}. Accordingly, the expectation in Eq.~\eqref{eq:optim} vanishes and the objective simply becomes $g(\set A';\theta_0)$, being $\theta_{0}$ the set of parameters representing perfect knowledge of the system.
Further, for this kind of attack, the objective function $g(\set A';\theta_0)$ can be defined as a distance function between the clustering result $\mathcal C$ obtained from the untainted data $\set D$ and the clustering result $\set C'=f_{\set D}(\set D')$ restricted to the same data (through a projection operator $f_{\set D}$), but obtained from the tainted data $\set D' = \set D\cup\set A'$ (\ie, including the set $\set A'$ of attack samples).
The objective can be thus written as $g(\set A';\theta_0)=d_\text{c}(\set C,f_{\set D}(\set D\cup\set A'))$, where $d_\text{c}$ is a suitable distance function between clusterings. 
Note that poisoning samples are excluded from the computation of the objective function since the attacker's goal is to maximally subvert the clustering output on the \emph{untainted} input data, and not on the poisoning samples (which may otherwise bias the evaluation of the attack's impact).

If the clustering algorithm $f$ assigns each sample to a cluster, the clustering result $\set C$ can be  represented as a matrix $\mat Y\in \{0, 1 \}^{\con n\times\con k}$ ($k$ being the number of clusters found), where each $(i,j)^{\rm th}$ component equals $1$ if the $i^{\rm th}$ sample is assigned to the $j^{\rm th}$ cluster, and $0$ otherwise. Within this setting, a possible distance function between clusterings amounts to counting how many pairs of samples have been clustered together in one clustering and not in the other, or viceversa:
\begin{equation}
	d_\text{c}(\mat Y,\mat Y')=\Vert \mat Y\mat Y\T-\mat Y'{\mat Y'}\T\Vert_F\,,
	\label{eq:distance}
\end{equation}
where $\Vert\cdot\Vert_F$ is the Frobenius norm, and each element of the matrix $\mat Y\mat Y\T \in \{0,1\}^{\con n\times\con n}$ (and, similarly, of $\mat Y'{\mat Y'}\T$) represents whether the corresponding pair of samples has been clustered together (1) or not (0).

As mentioned earlier, to poison the clustering process the adversary can add a set $\set A'$ of attack samples to the input data $\set D$. We bound the adversary's capability here by limiting the maximum number of injected poisoning samples to $\con m$, \ie{} $|\set A'| \leq \con m$. 
Additional constraints on the set of attack samples can be identified depending on the given feature representation, to facilitate the fabrication of real samples exhibiting the desired feature values.
In general, we denote the set of constraints to be fulfilled by the poisoning attack as $\set A' \in \Omega_{p}$. 
To give a concrete example, consider that \Malheur  can be configured to extract binary feature vectors that are subsequently normalized to have unitary $\ell_{2}$-norm.
In this case, the set of constrained attack samples can be expressed as:
\begin{equation}
\Omega_{p} =\left\{ \{\vct a'_i\}_{i=1}^{\con m}\,:\,  \vct a'_{i} \in \{0, 1/||\vct a'_{i}||_{2}\}^{\con d}  \text{ for }i=1,\cdots, \con m \right\} \, ,
\label{eq:malheur-constraint}
\end{equation}
where $\con d$ is the number of features, and $|| \cdot ||_{2}$ denotes the $\ell_{2}$-norm of a vector.

In general, the optimal attack strategy for poisoning attacks with perfect knowledge can be therefore derived from Eq.~\eqref{eq:optim} and written independently from the specific clustering algorithm as:
\begin{equation}
	\displaystyle
	\begin{array}{rl}
		\text{maximize}&d_\text{c}(\set C,f_{\set D}(\set D\cup\set A')) \\
		\text{s.t.}&\set A'\in\Omega_\text{p}\,. 
	\end{array}
	\label{eq:optim_pois}
\end{equation}

Unfortunately, this problem can not be solved analytically only if the
clustering output is analytically predictable, which is not usually
the case. We have thus to resort to suitable heuristics depending on
the considered clustering algorithm to devise effective attacks. In
the next section we investigate heuristics to solve the above problem
\citep[see][]{biggio13-aisec,biggio14-spr} and poison the single-linkage
hierarchical clustering algorithm, as we will exploit them in our case
study against \Malheur .

\subsection{Poisoning single-linkage hierarchical\\clustering}
\label{sect:attacking-hierarchical}

Before describing the heuristics for poisoning the single-linkage clustering algorithm, it is worth pointing out that this algorithm, as any other variant of hierarchical clustering, outputs a \emph{hierarchy} of clusterings~\cite{Jain99}.
Such a hierarchy is constructed by initially considering each data point as a single cluster, and  iteratively merging the closest clusters together, until a single cluster containing all data points is obtained. 
Clusters are merged according to a given distance measure, also referred to as \emph{linkage} criterion.
In the \emph{single-linkage} variant, the distance between any two clusters ($\set C_{i}$, $\set C_{j}$) is defined as the minimum Euclidean distance between all possible pairs of samples in $\set C_{i} \times \set C_{j}$.

To obtain a given data partitioning into clusters, a suitable cutoff distance has to be chosen. This determines the maximum intra-cluster distance for each cluster, and, thus, indirectly, the total number of clusters.
We follow the approach of Biggio et al.~\cite{biggio13-aisec,biggio14-spr} and select the cutoff distance that achieves the minimum distance between the clustering obtained in the absence of attack $\set C$ and the one obtained in the presence of poisoning, \ie, $\min d_c(\set C,f_{\set D}(\set D\cup\set A'))$. The reason is that this is the worst-case cutoff criterion for the attack, which is thus expected to work potentially even better under less pessimistic choices of the cutoff distance. 

Given a suitable criterion for selecting the cutoff distance, it is possible to model the clustering output as a binary matrix $\mat Y\in\{0,1\}^{\con n\times\con k}$ indicating the sample-to-cluster assignments, and thus use the distance measure $d_{c}$ defined in Eq.~\eqref{eq:distance} as the objective function in Problem~\eqref{eq:optim_pois}. This problem has then been solved by means of specialized search heuristics specifically tailored to the considered clustering algorithm.
In particular, we have considered greedy optimization approaches in which the attacker aims to maximize the objective function by adding one attack sample at a time, \ie, $|\set A^{\prime}|=\con m=1$.
We have found that the objective function is often maximized when the attack point is added in between clusters that are sufficiently \emph{close} to each other. The reason is that such an attack tends to decrease the distance between the two clusters, thus causing the algorithm to potentially merge them into a single cluster.

\begin{figure*}[t]
\centering
\includegraphics[width=1.2\textwidth,  trim=10 450 200 0, clip]{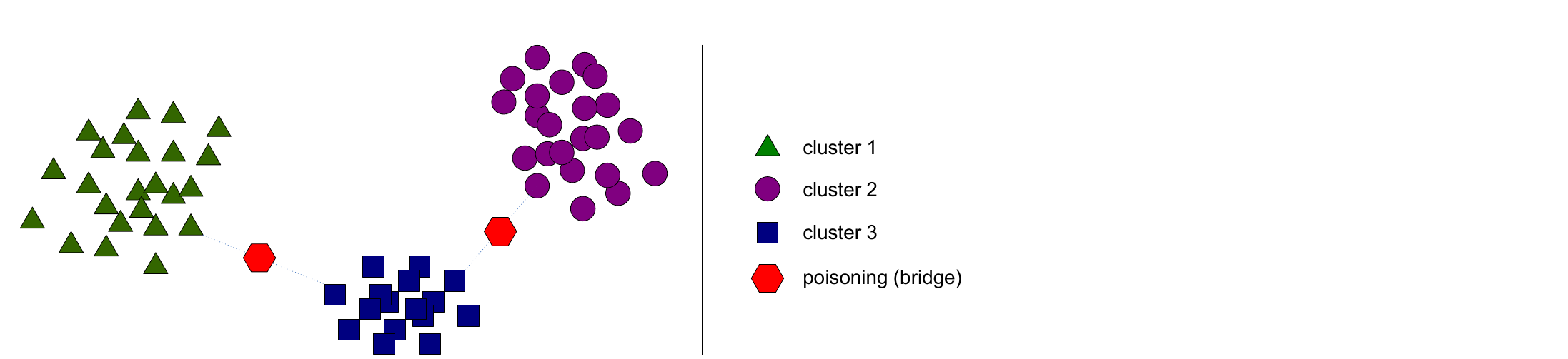}
\vspace{-5pt}
\caption{Bridge-based attacks against single-linkage clustering. The candidate attack samples connecting the $k-1$ closest clusters are highlighted as red hexagons.}
\label{fig:example}
\end{figure*}

\ourparagraph{Bridge-based attacks} Based on this observation, we have thus devised a family of attacks that aim to iteratively \emph{bridge} the closest clusters.
Let us assume that at each iteration we are given a set of $k$ clusters, and we have to select the best attack point to be added to the current dataset. Each bridge-based attack generates the same set of $k-1$ candidate attack points, by considering the $k-1$ links between pairs of points that have been cut to separate the current clustering from the top of the hierarchy, \ie, the $k-1$ shortest connections between clusters.
Each candidate attack point is then computed as the midpoint between the points in each of the $k-1$ identified pairs, as conceptually represented in Fig.~\ref{fig:example}.
The difference among the bridge-based attacks relies only on how the best attack point is selected at each iteration. 

\ourparagraph{Bridge (Best)} This strategy adds each candidate attack point to the current dataset, one at a time, re-runs the clustering algorithm on such data, and chooses the attack point that maximally increases the objective function. 
This is clearly a computationally-intensive procedure, especially for large datasets. 

\ourparagraph{Bridge (Hard)} This strategy aims to improve efficiency by avoiding us to re-run the clustering $k$ times at each attack iteration. The underlying idea is to approximate the clustering result $\mat Y^{\prime}$ on the current dataset including the considered candidate attack point, without re-computing the clustering explicitly. To this end, the attack point is assumed to effectively merge the two adjacent clusters.
For each point belonging to one of the two adjacent clusters, we thus set to $1$ ($0$) the value of $\mat Y^{\prime}$ corresponding to the first (second) cluster. This amounts to considering \emph{hard} clustering assignments. Once the estimated $\mat Y^{\prime}$ is computed, we evaluate the objective function using the estimated $\mat Y^{\prime}$, and select the attack point that maximizes its value. 

\ourparagraph{Bridge (Soft)} This is a variant of the latter approach that estimates $\mat Y^{\prime}$ using soft clustering assignments instead of hard ones.
In particular, the $(i,k)^{\rm th}$ element of $\mat Y^{\prime}$ is estimated as the posterior probability that the $i^{\rm th}$ sample belongs to the $k^{\rm th}$ cluster, using a Gaussian Kernel Density Estimator (KDE) with bandwidth parameter $h$. When $h$ is too small, the posterior estimates tend to the value of $1/k$, \ie, each point is assigned to any cluster with the same probability. When $h$ is too high, instead, they tend to hard assignments. As a rule of thumb, the value of $h$ should be thus comparable to the average distance between all possible pairs of samples in the dataset.
The rationale of this strategy is to try finding connections that can potentially merge large clusters with more than one attack sample, to mitigate the limitation of our greedy approach.

\section{A Case Study: Malheur}
\label{sect:malheur}

To illustrate the effect of the proposed poisoning attacks in a
practical setting, we conduct a case study with the open-source tool
\Malheur .\footnote{\url{http://www.mlsec.org/malheur}} The tool
implements techniques for clustering and classification of program
behavior and has been applied in different settings for analyzing
malware in the wild \cite{BruEpaHof+10,rieck11-jcs,HuShi13}. The
analysis realized by \Malheur builds on four basic steps.

\begin{enumerate}

\item \emph{MIST Representation.} As the first step, the behavior of
  malware binary is monitored in a sandbox environment and stored as
  \emph{MIST reports}~\cite{TriWilHolRie10}.  In this format, the
  behavior of a program is described as a sequence of events, where
  individual execution flows of threads and processes are grouped in a
  single, sequential report. Each event encodes one monitored system
  call and its arguments, where the arguments are arranged in
  different levels of blocks, reflecting behavior with different
  degree of granularity. Depending on the configuration of \Malheur,
  the monitored behavior can be analyzed at these different \emph{MIST
    levels} \cite{rieck11-jcs}.

\item \emph{Embedding.} As the next step, \Malheur embeds the
  monitored behavior in a high-dimensional vector space, where each
  dimension is associated with a short sequence of $q$ events---a so
  called \emph{$q$-gram}.  If a $q$-gram occurs in the monitored
  events of a program, the respective dimension is set to 1 in its
  vector, otherwise it is set to 0.  To enable a fair
  comparison of programs that strongly differ in the amount of
  observed events, each vector $\vct x$ is additionally normalized, such
  that $||\vct x||_2 = 1$, namely, projecting the vectors onto a
  hypersphere of unit radius in the vector space.

\item \emph{Clustering.} For partitioning the embedded behavior into
  groups, \Malheur implements an efficient variant of hierarchical
  clustering that supports single-linkage and complete-linkage
  hierarchical clustering.  To alleviate the quadratic run-time
  complexity of these clustering algorithms, the tool can approximate
  the underlying data by limiting the analysis to a small subset of
  prototypes. For our case study, we disable this functionality and
  instead apply \Malheur without prototype-based approximation.

\item \emph{Classification.} Finally, \Malheur  supports assigning
  unknown behavior to previously discovered clusters. This assignment
  is realized using a nearest-neighbor classification, where a new
  vector is assigned to a nearby cluster if it appears within a
  certain distance to its members. This nearest-neighbor
  classification can be approximated by searching for nearest
  neighbors in a set of prototypes instead of all cluster members.  We
  again disable this functionality and operate on the full data for
  our case study.
\end{enumerate}

Each of the four steps supports different parameters that can be
adapted in the configuration of \Malheur.
For our case study, we start with a basic setup by using MIST
level 1, setting the q-gram length to 1 and especially considering
single-linkage clustering by disabling the prototype-based approximation
 used by \Malheur{}. The use of the latter would indeed imply a sort of
complete-linkage pre-processing clustering step,
which would in turn require us to significantly revisit the derivation of a proper poisoning attack. We therefore leave this issue to future work.
Finally, although this setup slightly simplifies our attack, previous work has already shown that creating artificial $q$-grams of system calls, and similarly
the use of different MIST levels, is not a challenge for an attacker
\cite{TanMax02,TanKilMax02,WagSot02}, especially if she has full control over the behavior, as in the case of malware.

\section{Experimental Evaluation}
\label{sect:evaluation}

In this section we apply the aforementioned evaluation framework to a concrete case study: we evaluate the worst-case effects of the poisoning attacks described in Section \ref{sect:attacking-hierarchical} using \emph{real} malware samples, and against a real-world tool for behavioral malware clustering. 
In Section~\ref{sect:datasets} we present the datasets employed for our investigation. Then, in Section~\ref{sect:setup} we provide all relevant details about the experimental setup and evaluation metrics. In Section~\ref{sect:attack-strategies} we summarize the main attack strategies implemented for the evaluation, including the modifications to the derivation of poisoning attacks that allow us to deal with the specific feature representation exploited by \Malheur{}. Finally, in Section~\ref{sect:results} we present and discuss our experimental results.

\subsection{Datasets}
\label{sect:datasets}

For our experiments and evaluation we make use of two different
datasets: first, the data that was originally considered by Rieck \etal
in~\cite{rieck11-jcs}, and second, a dataset consisting of more recent malware samples 
collected in 2013.

\ourparagraph{Malheur data}
This dataset consists of a selection of 3131 malware samples collected in a
period of 3 years up to August 2009, and made publicly available in the
same year.\footnote{http://pi1.informatik.uni-mannheim.de/malheur/} It
comprises a \textit{reference dataset} that was used to calibrate the
clustering algorithm in~\cite{rieck11-jcs}, and 7 \textit{application
datasets} for evaluating and testing their approach. The latter
represent malware found on the Internet within 24 hours on 7 consecutive
days.
For our experiments we stick to a similar setup in order to ensure the comparability with the original approach and optimally show the practicality of our attack.

\ourparagraph{Recent Malware data}
In addition to the data used for the \Malheur project, we
gathered malware samples from most prominent families in 2013.
Similarly to~\cite{rieck11-jcs} we rely on the popular antivirus
scanner by Kaspersky Lab for this ranking and labeling of the malware
samples. We chose 5 of the top 10 detections according to a recent
threat report~\cite{Kas14}, and selected those families for
which we were able to gather more than 100 but at most 150 samples. A
summary of the exact numbers is given in Table~\ref{tab:dataset2}.

\begin{table}[t]
  \centering
  \begin{tabular}{l c }
    \toprule
    \bf Dataset & \bf ~~Number of samples \\
    \midrule

    \textit{DangerousObject.Multi.Generic} & 129 \\
    \textit{Trojan.Win32.Generic}          & 120 \\
    \textit{Virus.Win32.Sality.gen}        & 112 \\
    \textit{Trojan.Win32.Starter.lgb}      & 150 \\
    \textit{Virus.Win32.Nimnul.a}          & 146 \\
    \midrule
    Total                                  & 657 \\
    \bottomrule \\
  \end{tabular}
  \vspace{-10pt}
  \caption{\label{tab:dataset2} Summary of the malware families collected in 2013 for the \emph{Recent Malware} data.}
\end{table}

\smallskip
When running \Malheur  on the aforementioned datasets using the MIST-level-1 binary embedding discussed in Sect.~\ref{sect:malheur}, we have respectively found 85 and 78 distinct feature values (\ie, $1$-grams).

\subsection{Experimental Setup}
\label{sect:setup}

To fairly evaluate the clustering process, we randomly split each dataset into two disjunct portions of equal size, namely, $T$ and $S$. The $T$ portion is used to calibrate the clustering algorithm, and, in particular, to estimate the cutoff distance (see Sect.~\ref{sect:attacking-hierarchical}). As suggested in~\cite{rieck11-jcs}, we select as the optimal cutoff distance the one that maximizes the F-measure (see below for its definition).
The $S$ split is then used to evaluate the calibrated clustering on unseen malware against an increasing percentage of poisoning attacks.
This procedure is repeated five times and results are averaged over these repetitions.
In our experiments, the value of the cutoff distance has been found to be 0.49 on the \emph{Malheur} dataset and 0.63 on the \emph{Recent Malware} dataset, on average, with a negligible standard deviation in both cases.
Although in these experiments we assume that the cutoff distance is known to the attacker, more realistically an attacker can try estimating it from the data in a more conservative manner (\ie, essentially underestimating its real value), eventually poisoning the clustering result at the expense of using more poisoning samples.
Clustering results are evaluated according to the three measures given below.
\begin{enumerate}
\item The objective function in Eq.~\eqref{eq:optim_pois}, with $d_{c}$ given by Eq.~\eqref{eq:distance}, that measures the distance of the current clustering from that obtained in absence of poisoning. 
\item The number of clusters, that helps us to gain a better understanding of how the attacks taint the clustering process. In particular, since the considered poisoning attacks are expected to ``bridge'' clusters, the number of clusters should decrease as the attack progresses. 
\item The F-measure~\cite{Rijsbergen79,rieck11-jcs}, defined as the harmonic mean between precision ($\pi$) and recall ($\rho$), \ie, $2\frac{\pi \rho}{\pi+\rho}$. The latter are respectively computed as $\pi = \frac{1}{n} \sum_{j} \max_{i} \mat c_{ij}$, and $\rho = \frac{1}{n} \sum_{i} \max_{j} \mat c_{ij}$, where $\mat c_{ij}$ is the number of malware samples belonging to the $i$-th family present in the $j$-th cluster~\cite{rieck11-jcs}. Precision reflects how well individual clusters agree with malware families, whereas recall measures to which extent malware families are scattered across clusters. Both measures provide complementary information about the quality of clustering results, summarized by the F-measure. 
\end{enumerate}

\subsection{Attack Strategies}
\label{sect:attack-strategies}

\begin{figure}[t]
\centering
\includegraphics[width=0.16\textwidth ]{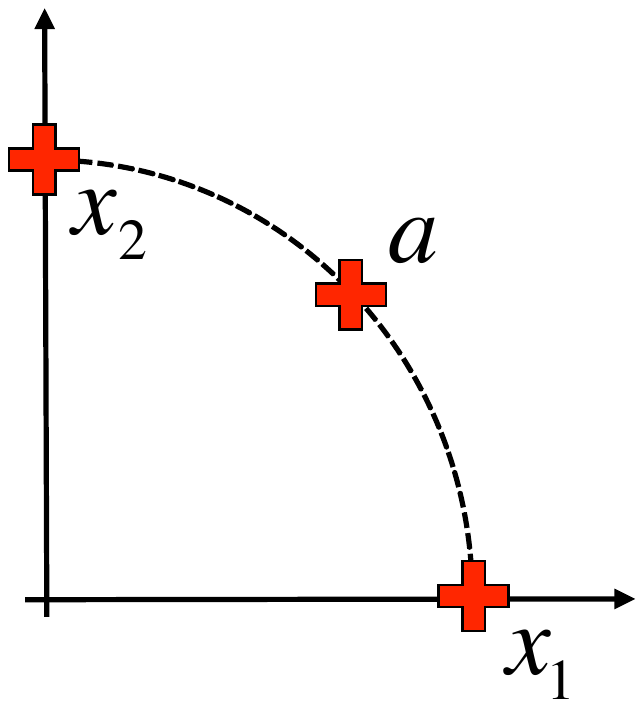}
\vspace{-5pt}
\caption{Computation of a bridge-based attack against single-linkage clustering, using the feature representation of \Malheur, \ie, a binary embedding with points additionally  projected onto a unit hypersphere (\ie, with unit $\ell_{2}$ norm). This example considers a simple two-dimensional feature set, where we highlighted the only three admissible points, \ie, the samples $\vct x_{1} = (0,1)$ and $\vct x_{2} = (1,0)$, and their ideal bridging point $\vct a = (1/\sqrt 2,1/\sqrt 2)$. Besides this simple case, the creation of effective bridge-based attacks in this space is generally much more challenging than that considered in our previous work~\cite{biggio14-spr,biggio13-aisec}, due to the restrictions imposed by the given feature representation.} 
\label{fig:bridge-malheur}
\end{figure}

In this section we explain how we generate the poisoning samples to attack \Malheur. 
First, as \Malheur is configured with binary embedding and $\ell_{2}$ normalization in our case (see Sect.~\ref{sect:malheur}), the feature vectors of poisoning samples have to fulfill the constraints given by the set $\Omega_{p}$ in Eq.~\eqref{eq:malheur-constraint}.
Besides these constraints, another fundamental pre-requisite that we impose is that poisoning points have to represent realistic malware samples. The reason is that a sample that does not exhibit any malicious or intrusive functionality may not be included into the set of collected malware to cluster, and the attack would be trivially defeated.\footnote{Note however that the main goal of poisoning samples is not to preserve the malicious functionality of the embedded malware code, which is required here only to avoid having such samples discarded by a simple preliminary antivirus analysis. Instead, their primary goal is to subvert the clustering output on the rest of the data, in order to produce a less effective characterization of malware families. This may indeed not only lead to lower malware detection rates or higher false alarm rates, but it also makes more difficult to identify the proper countermeasures or removal tools in case of infection.} 
Therefore, we generate every poisoning sample by first selecting a malware sample from the given $S$ split, and then manipulating its features by only increasing their value.
Note that the value of a feature can only be increased from 0 to $1/||\vct a||_{2}$, being $||\vct a||_{2}$ the $\ell_{2}$-norm of the attack sample. We thus refer in the following to this kind of manipulation as feature addition, for short. This manipulation indeed preserves the malicious functionality of the initially-selected malware, as it does not compromise the set of instructions required to execute the original malicious code.
Moreover, before adding any candidate poisoning point to the data, we verify whether another point with the same feature values is already present. If this is the case, we discard the current sample and choose the next best candidate attack point. This allows us to discard duplicate attack points, as their presence may worsen the attack progress.

An important consequence of the particular embedding used by \Malheur  is that it affects the way we compute the \emph{bridge} between any two points to create our candidate attack samples. This is a rather important distinction with respect to our previous work in \cite{biggio14-spr,biggio13-aisec}.
In fact, the midpoint in this case can not be computed as the average of the two neighboring points, as it is instead possible, for instance, when real-valued features are used. However, the point that is as equidistant as possible from each of the two neighboring points can be found by cloning the neighboring point with the smaller norm first, and then starting adding features to it that are not null in the other neighboring point, until the candidate attack point is as equidistant as possible from the two points.
A simple two-dimensional example is given in Fig.~\ref{fig:bridge-malheur}. The only drawback of this procedure is that the candidate attack point may be sometimes farther from the two neighboring points than they are with respect to each other. In these cases, the attack may not be effective, as it may not effectively bridge the two neighboring clusters. 

In these experiments we consider six distinct poisoning attack strategies. In addition to the three bridge-based attacks defined in Sect.~\ref{sect:attacking-hierarchical}, we consider \textit{Random} and \textit{Random (Best)} as in \cite{biggio13-aisec}, and a variant of our bridge-based attacks named \textit{F-measure (Best)}.  \textit{Random} generates any attack point by cloning a randomly-selected malware from the available set $S$, and adding to it a random number of features. \textit{Random (Best)} works similarly, with the difference that not one but $k-1$ attack points are selected at random, being $k$ the actual number of clusters at any given attack iteration. Then, the objective function is evaluated for each candidate point by re-running the clustering algorithm, and the best attack point is chosen. 
\textit{F-measure (Best)} works as \textit{Bridge (Best)}, but chooses the best candidate attack point as the one that minimizes the F-measure instead of maximizing the objective function $d_{c}(\mat Y, \mat Y')$. As \emph{Random (Best)} and \textit{Bridge (Best)}, this strategy also requires evaluating the clustering result $k-1$ times to determine the best attack at each iteration, while the other strategies are computationally lighter. 
As for \textit{Bridge (Soft)}, we set the kernel bandwidth $h$ as the average distance between each possible pair of samples in the data, which yielded $h \approx 0.2$ in each run.
We finally point out that, if more than one candidate attack point exhibit the same value of the desired function (either the objective function or the F-measure, depending on the attack strategy), we select the one that produces the smaller number of clusters. If the tie persists, we break it at random.

\subsection{Results}
\label{sect:results}
\begin{figure*}[ht]
\centering
\includegraphics[width=0.82\textwidth]{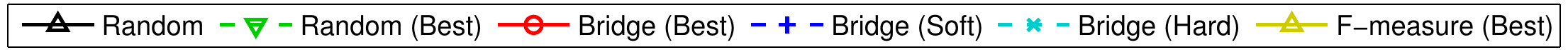}\\
\includegraphics[width=0.4\textwidth]{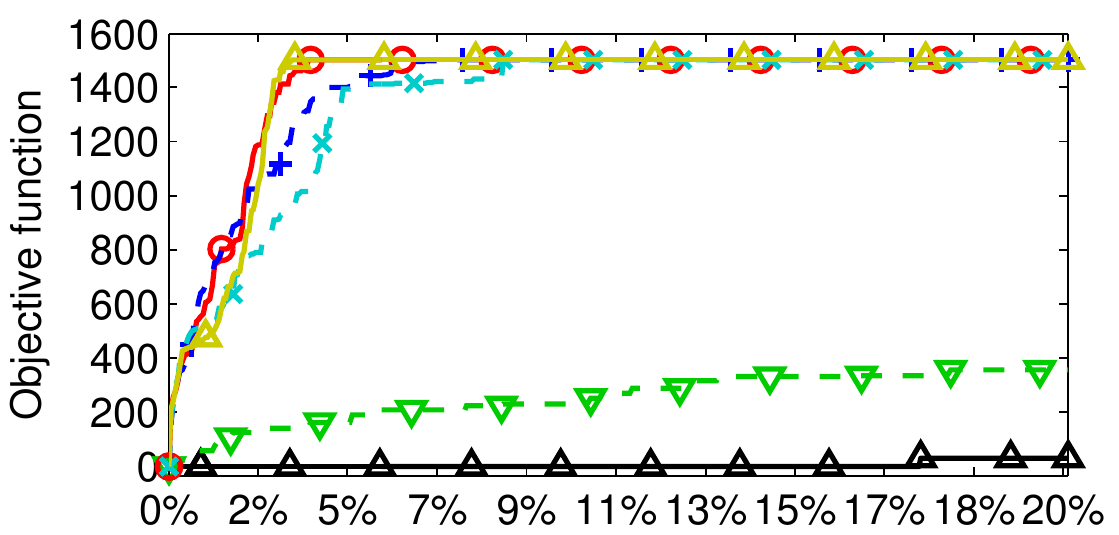}
\includegraphics[width=0.4\textwidth]{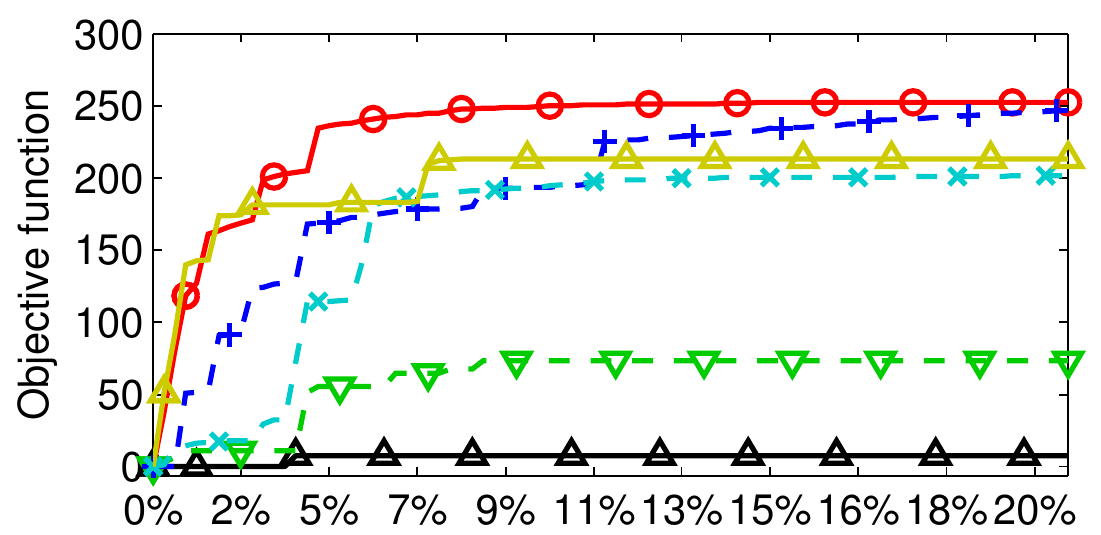}\\
\includegraphics[width=0.4\textwidth]{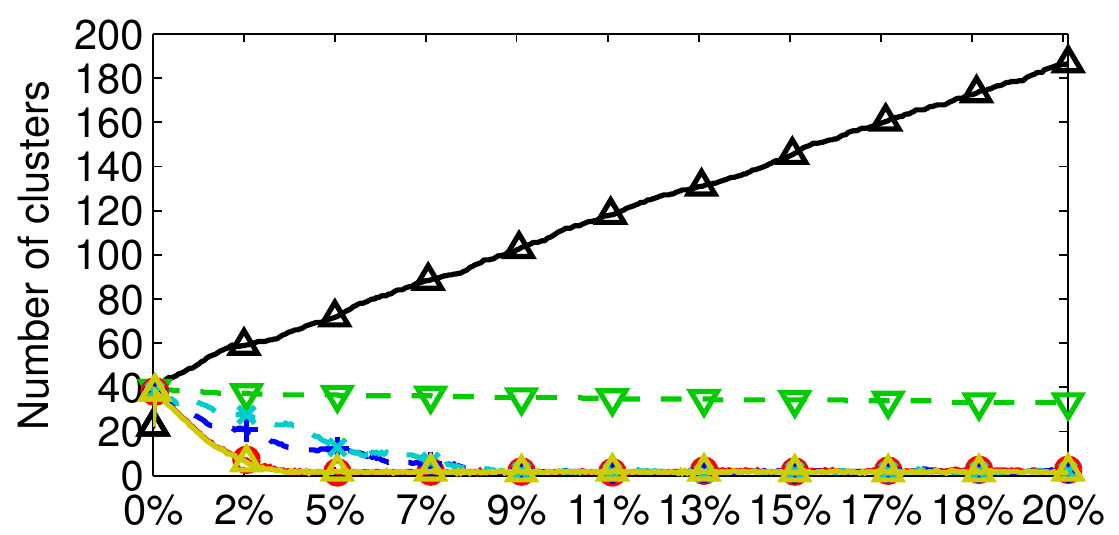}
\includegraphics[width=0.4\textwidth]{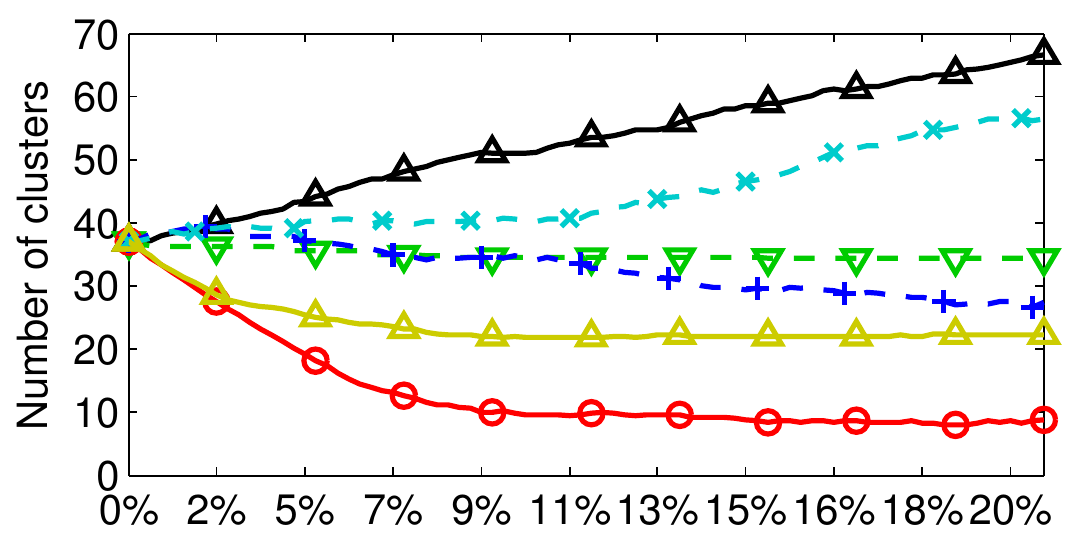}\\
\includegraphics[width=0.4\textwidth]{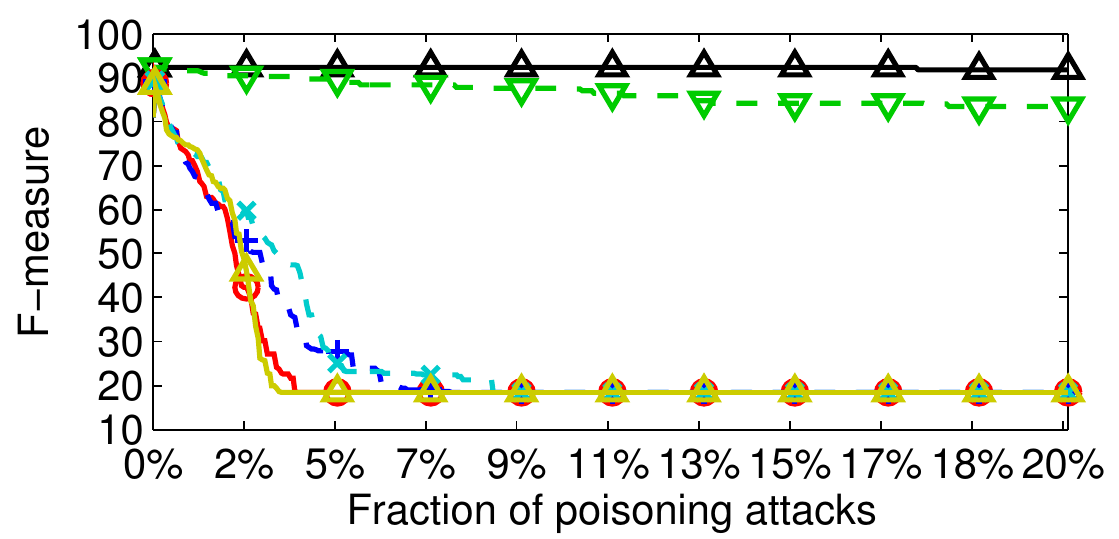}
\includegraphics[width=0.4\textwidth]{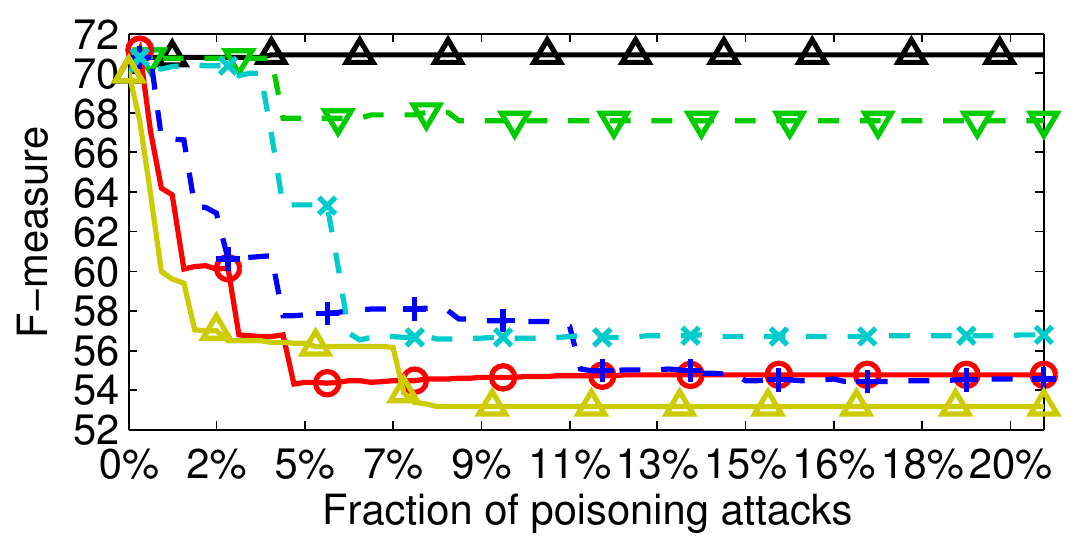}
\caption{Results for the \emph{Malheur} dataset (left column) and the \emph{Recent Malware} dataset (right column).}
\label{fig:results}
\end{figure*}

Results for the \emph{Malheur} and the \emph{Recent Malware} datasets are presented in Fig.~\ref{fig:results}. For each dataset, we show how the value of the objective function, the number of clusters, and the F-measure change for an increasing percentage of injected poisoning samples.  
We observe a similar behavior of these metrics for both datasets, which is  summarized below in two points.
\begin{enumerate}
\item Simply injecting random points does not allow one to significantly worsen the quality of the resulting clustering. We can in fact observe that, for the \textit{Random} attack, neither the value of the objective function nor the F-measure are affected at all. The reason is that each of the randomly-generated attack points is too far from the other clusters and it is thus clustered as a singleton, without affecting the clustering result on the rest of the samples. \emph{Random (Best)} performs slightly better, as it clearly makes $k-1$ attempts at each iteration to find a better attack point, instead of one.
\item Maximizing the considered objective function actually allows us to reduce the number of clusters, and, thus, to compromise the quality of the resulting clustering, despite it does not incorporate any knowledge of the problem domain,  of the clustering algorithm, and of the features used. Furthermore, looking at Fig.~\ref{fig:results}, we can observe that the bridge-based strategies that maximize the objective function achieve similar performances to \emph{F-measure (Best)}, which instead minimizes the F-measure. Whereas the objective function is general, the F-measure takes into account the ground truth of the problem. We can therefore reasonably argue that the proposed objective function and the consequent attack strategies can be successfully employed to attack also systems different from \Malheur. 
\end{enumerate}

Some further comments can be made separately for the two data sets. 
What appears evident from the results on the \emph{Malheur} dataset is that injecting an even small percentage of poisoning points reduces significantly the number of clusters.  \textit{Bridge (Best)} and  \textit{F-measure (Best)} are able to reduce the number of cluster from an initial value of 40 to a value of 5 with only the 2\% of injected samples. If we further increase such percentage up to 5\% a single, large cluster is created, where all the initial ones are merged. \textit{Bridge (Soft)} and \textit{Bridge (Hard)}  appear to be a bit less effective since they require a slightly higher percentage of injected samples to achieve similar results. Nevertheless, it is worth pointing out that, from a computational standpoint, both these strategies are significantly less expensive than the \textit{Best} strategies. 

On the \emph{Recent Malware} dataset the considered attacks appear to be less effective.
In particular, the bridge-based attacks here are not able to merge all the clusters into a unique cluster. At some point, instead, it happens that the strategies are no longer able to inflict any damage to the current clustering. The reason is that the candidate bridge points in this case are selected too far from their corresponding neighboring points, and the former are thus clustered apart instead of successfully merging the desired clusters.
We argue that this may be somehow due to the smaller number of features found in this dataset, as this factor limits the number of manipulations that the attacker can make to find a suitable attack point.
This may be an interesting starting point for future work to understand how to improve robustness of clustering algorithms to poisoning attacks by restricting the feature set and the number of potential manipulation the attacker can make on the attack samples.
Nevertheless, one should keep in mind that, in this case, the objective function reaches anyway the value of 250 for \emph{Bridge (Best)}, which still means that 250 pair of samples out of 329 samples have changed their clustering assignment with respect to the clustering in the absence of poisoning.

\section{Conclusions and Future Work}
\label{sect:conclusions}

A widespread approach for coping with the plethora of novel malware
are clustering algorithms from the area of machine learning.  While
these algorithms can help grouping similar malware samples
automatically, they have not been originally designed to operate in an
adversarial setting. Our work shows that, by leveraging on
vulnerabilities of clustering algorithms, an attacker can
significantly impact the performance of malware clustering. In our
evaluation, only a small fraction of poisoning samples is necessary to
largely destroy the recovery of families in a dataset of real malware.
In particular, in this work we have considered \Malheur,
\ie, a popular malware clustering
tool. We have modified previously-proposed poisoning attacks to cope with
its specific feature representation, and to incorporate the corresponding application-specific constraints in the creation of real, poisoning malware samples.
Although we have focused on a particular setup of this tool,
we argue that attacks to other setups and clustering systems should not be considered a major challenge for a sophisticated attacker. Creating behavioral features artificially may
be more or less difficult depending on the underlying sandbox
environment, yet the exploited vulnerability resides in the clustering
algorithms and thus can hardly be fixed by changing the feature
representation. As a result, our work casts serious doubt about the
security of \emph{some} clustering algorithms in malware analysis systems,
and there may be considerable need for novel algorithms that are more robust against
poisoning and malicious noise.

Future extensions of this work may include:
investigation of attacks in which the adversary has only limited knowledge of the system, \ie, attacks in which the input data is not known to the attacker, who may realistically only collect surrogate data from the same sources;
development of poisoning attacks that may target a larger family of clustering algorithms (instead of considering only specialized heuristics); and development of appropriate countermeasures to improve security of clustering algorithms against adversarial threats and well-crafted attacks.

It is also worth remarking here that poisoning attacks are not the only kind of attack that may be incurred by a clustering-based system operating in an adversarial setting;
\eg, if some of the clusters are used to characterize the behavior of legitimate users or software, an attacker may aim to manipulate the malware behavior in order to mimic the legitimate samples, without significantly altering the clustering output on the rest of the data. This attack has been referred to as \emph{obfuscation} attack in \cite{biggio13-aisec}. We refer the reader to the same work for a detailed taxonomy of potential attacks against clustering. However, the implementation of such attacks for more realistic application scenarios and specific feature representations remains a non-trivial open issue, which should be addressed as done in this paper for poisoning attacks and behavioral malware clustering.


\section{Acknowledgments} We thank Cristian Milia for supporting
during tests. This work has been partly supported by the Regional
Administration of Sardinia (RAS), Italy, within the projects
``Security of pattern recognition systems in future internet''
(CRP-18293), and ``Advanced and secure sharing of multimedia data over
social networks in the future Internet'' (CRP-17555). Both projects
are funded within the framework of the regional law \emph{L.R. 7/2007,
  Bando 2009}. Furthermore, we acknowledge funding from BMBF under the
project PROSEC (FKZ 01BY1145).  The opinions, findings and conclusions
expressed in this paper are solely those of the authors and do not
necessarily reflect the opinions of any sponsor.


%
%
%

\renewcommand\bibname{References}
\makeatletter
\renewcommand\bibsection%
{
  \section{REFERENCES
    }
}
\makeatother
\makeatletter
\renewcommand\@biblabel[1]{#1.}
\makeatother

\balance
\bibliographystyle{abbrv}
\bibliography{bibDB}

\end{document}